\documentclass[letterpaper, 10 pt, conference]{ieeeconf}  

\IEEEoverridecommandlockouts                              

\usepackage{microtype}
\usepackage{graphicx}
\usepackage{subfigure}
\usepackage{booktabs} 
\usepackage{balance}
\usepackage{makecell}
\usepackage{bm}
\makeatletter
\let\NAT@parse\undefined
\makeatother
\usepackage{hyperref}
\usepackage[numbers,sort&compress]{natbib}

\usepackage{amssymb}
\usepackage{multirow}
\usepackage{amsmath}
\usepackage{algorithm,algpseudocode}
\usepackage{wrapfig}
\usepackage{color}
\usepackage{soul}
\usepackage{xspace}
\usepackage{cleveref}

\Crefname{figure}{Figure}{Figures}
\crefname{figure}{Figure}{Figures}
\crefname{table}{Table}{Tables}
\crefformat{table}{Table~#1}
\crefformat{equation}{eqn.(#1)}
\Crefformat{equation}{Eqn.(#1)}
\crefformat{algorithm}{Algorithm~#1}

\newcommand{\bea}{\begin{eqnarray}}
\newcommand{\eea}{\end{eqnarray}}
\newcommand{\beas}{\begin{eqnarray*}}
\newcommand{\eeas}{\end{eqnarray*}}
\newcommand{\leftm}{\left[\begin{array}}
\newcommand{\rightm}{\end{array}\right]}

\usepackage{diagbox,hhline}

\def\span{{\mbox{\bf span}}}

\newcommand{\red}[1]{{\color{red}#1}}
\newcommand{\blue}[1]{{\color{blue}#1}}
\definecolor{commentclr}{RGB}{110, 149, 204}

\usepackage{stackengine}
\stackMath

\newif\ifcomments
\commentstrue

\usepackage{color}
\usepackage{soul}
\usepackage{xspace}
\usepackage{cleveref}

\Crefname{figure}{Figure}{Figures}
\crefname{figure}{Figure}{Figures}
\crefname{table}{Table}{Tables}
\crefformat{table}{Table~#1}
\crefformat{equation}{eqn.(#1)}
\Crefformat{equation}{Eqn.(#1)}
\crefformat{algorithm}{Algorithm~#1}

\newif\ifcomments
\commentsfalse 
\commentstrue

\ifcomments
	\newcommand{\dXX}[1]{\color{red}DK: (#1)\color{black}\xspace}  
	\newcommand{\cXX}[1]{\color{blue}CS: (#1)\color{black}\xspace}  
	\newcommand{\XX}[1]{\color{green}JH: (#1)\color{black}\xspace}  
\else
    \newcommand{\dXX}[1]{}  
	\newcommand{\cXX}[1]{}  
	\newcommand{\XX}[1]{}  
\fi

\usepackage{caption}

\title{\LARGE \bf
ROMAX: Certifiably Robust Deep Multiagent Reinforcement Learning via Convex Relaxation
}

\author{Chuangchuang Sun$^{1}$ Dong-Ki Kim$^{1}$ and Jonathan P. How$^{1}$
\thanks{$^{1}$Laboratory for Information \& Decision Systems (LIDS), Massachusetts Institute of Technology, 77 Massachusetts Ave, Cambridge, MA 02139. 
{\{\tt\small ccsun1,dkkim93,jhow\}@mit.edu}}%
}

\begin{document}

\maketitle
\thispagestyle{empty}
\pagestyle{empty}

\begin{abstract}
In a multirobot system, a number of cyber-physical attacks (e.g., communication hijack, observation perturbations) can challenge the robustness of agents. This robustness issue worsens in multiagent reinforcement learning because there exists the non-stationarity of the environment caused by simultaneously learning agents whose changing policies affect the transition and reward functions. In this paper, we propose a minimax MARL approach to infer the worst-case policy update of other agents. As the minimax formulation is computationally intractable to solve, we apply the convex relaxation of neural networks to solve the inner minimization problem. Such convex relaxation enables robustness in interacting with peer agents that may have significantly different behaviors and also achieves a certified bound of the original optimization problem. We evaluate our approach on multiple mixed cooperative-competitive tasks and show that our method outperforms the previous state of the art approaches on this topic.
\end{abstract}

\section{Introduction}\label{sectioin:introduction}

Multirobot systems have recently attracted much attention in robotics. 
Compared to a single robot approach, a multirobot system provides several unique benefits, including 1) improved efficiency since a sophisticated problem can be decomposed into simpler sub-problems, distributed across robots, and then solved simultaneously and 2) improved mission success rate because a single robot failure can be addressed by another teammate~\cite{mellouli2007reorganization}.
These advantages have resulted in emerging multirobot applications, such as formation control~\cite{alonso2017multi}, cooperative manipulation~\cite{culbertson2021decentralized}, and human-swarm interaction~\cite{vasile2011integrating}. 

Multiagent reinforcement learning (MARL) provides a principled framework for solving problems in which multiple robots interact with one another in a shared environment. 
However, there remain difficulties in learning intelligent multiagent policies. 
Amongst these, instability in policy learning is particularly problematic in that agents generally show poor performance when interacting with unseen agents \cite{al-shedivat2018continuous,kim21metamapg}. 
While there are approaches that stabilize policy learning in MARL (e.g., centralized training and decentralized execution frameworks~\cite{lowe2017multi,foerster2018counterfactual}), agents generally overfit other agents' policies during training, resulting in a failure when playing against new strategies not interacted before. 
This robustness issue becomes more severe in a competitive setting, where an opponent can intentionally apply cyber-physical attacks (e.g., communication hijack, observation perturbations), fully exploit an agent's brittle policy, and thus dominate a game~\cite{gleave2019adversarial}.

\textbf{Our contributions.} To address the robustness problem, we propose a new framework, called \textbf{ROMAX}: \textbf{RO}bust \textbf{MA}RL via convex rela\textbf{X}ation. 
While the minimax optimization enables learning of robust multiagent policy~\cite{littman1994markov}, solving a general nonconvex-nonconcave minimax
optimization problem is computationally intractable~\cite{daskalakis2020complexity}. 
Assuming that each agent's policy is parameterized by deep neural networks, we develop a computationally efficient approach that can approximately solve the minimax optimization and infer the worst-case actions of other agents via the convex relaxation of neural networks. We note that this convex relaxation has an important benefit in that it can explore the approximate \textit{globally} worst situation while achieving \textit{certified} robustness from the guaranteed bound of the relaxation. 
We empirically evaluate our algorithm on multiple mixed cooperative-competitive tasks and show that ROMAX outperforms baselines by a significant margin, demonstrating the necessity to compute the worst-case scenarios to improve robustness in MARL.
\section{Related Works}\label{section:relate_works}
\textbf{Centralized training with decentralized execution.}
The standard approach for addressing non-stationarity in MARL is to consider information about other agents and reason about the effects of joint actions~\cite{hernandezLealK17survey}.
The recent studies regarding the centralized training with decentralized execution framework, for instance, account for the behaviors of others through a centralized critic~\cite{lowe2017multi,foerster2018counterfactual,yang18mean,wen2018probabilistic,kim20hmat}.
While this body of work partially alleviates non-stationarity, converged policies generally overfit the current behaviors of other agents and thus show poor performance when interacting with agents with new behaviors.
In contrast, our agents learn robust policies based on minimax optimization by applying convex relaxation. 


\textbf{Robust MARL.} 
Our framework is closely related to prior works that apply minimax optimization in multiagent learning settings~\cite{perolat17mean,graumoya18}.
Minimax provides a game-theoretical concept that encourages an agent to learn a robust policy by maximizing its performance in a worst-case scenario \cite{littman1994markov,osborne04minimax}.
One of the noticeable studies in this category is \cite{li2019robust}, which computes the worst-case perturbation by taking a single gradient descent step assuming that other agents act adversarial.
However, the single-step gradient approximation can only explore the locally worst situation and thus can still result in unstable learning. 
Our approach aims to address this drawback by computing the approximate globally worst situation based on convex relaxation. 
The work by \cite{lutjens2020certified} applies the similar linear relaxation technique in a single-agent robust RL problem to certify the robustness under uncertainties from the environments. 
However, in our multiagent settings, the robustness is more challenging to certify due to the concurrent policy learning amongst multiple agents.

\textbf{Ensemble training in MARL.}
Another relevant approach to learning a robust policy is ensemble training, where each agent interacts with a group of agents instead of a particular agent only \cite{lowe2017multi,Shen_How_2021,Schrittwieser2020MasteringAG}. 
For example, the population-based training technique, which was originally proposed to find a set of hyperparameters for optimizing a neural network \cite{Jaderberg17pbt}, was applied in MARL by evolving a population of agents \cite{Jaderberg_2019}. 
This approach showed robust and superhuman level performance in a competitive game. 
The literature on self-play, which plays against random old versions of itself to improve training stability and robustness, can also be classified into this category \cite{bansal2018emergent}. However, maintaining and/or evolving a population is often computationally heavy. Additionally, these methods do not employ minimax optimization, so agents may not be able to cope well with the worst scenario. 

\textbf{Learning aware MARL.}
Our framework is also related to prior works that consider the learning of other agents in the environment to address non-stationarity. 
These works include~\cite{zhang10lookahead} which attempted to discover the best response adaptation to the anticipated future policy of other agents. Our work is also related to \cite{foerster17lola,foerster2018dice} that shape the learning process of others.
Another relevant idea explored by~\cite{letcher2018stable} is to interpolate between the frameworks of~\cite{zhang10lookahead} and~\cite{foerster17lola} in a way that guarantees convergence while influencing the opponent's future policy. Recently, \cite{kim21metamapg} addresses non-stationarity by considering both an agent’s own non-stationary policy dynamics and the non-stationary policy dynamics of other agents within a meta-learning objective.
While these approaches alleviate non-stationarity by considering the others' learning, they do not solve the minimax objective and cannot guarantee robustness when playing against a new opponent. 
This weakness can be exploited by a carefully trained adversary agent~\cite{gleave2019adversarial}.

\textbf{Robustness verification and neural network relaxation.}
To verify the robustness of neural networks, it is important to compute the lower and upper bound of the output neurons under input perturbations.
In supervised learning settings, for example, the margin between predicting the ground-truth class and other classes indicates the robustness of neural networks (i.e., measuring the chance of misclassification). 
However, due to the nonconvexity in neural networks, the work by \cite{katz2017reluplex} proved that finding the true range of neural network's output is nonconvex and NP-complete. 
To address this issue, convex relaxation methods are proposed to efficiently compute the outer approximation (a more conservative estimate) of neural network's output range.
Many prior works are based on the linear relaxation of the nonlinear units in neural networks: FastLin~\cite{weng2018towards}, DeepZ~\cite{singh2018fast}, Neurify~\cite{wang2018efficient}, DeepPoly~\cite{singh2019boosting}, and CROWN~\cite{zhang2018efficient}. 
There are also other approaches based on semidefinite relaxation~\cite{raghunathan2018semidefinite,dvijotham2020efficient}, which admit tighter bounds but are more computationally expensive. 
See~\cite{salman2019convex} for in-depth surveys on this topic.
\section{Background}\label{section:Preliminary}
\subsection{Markov game}
Interactions between $n$ agents can be represented by a partially observable Markov game~\citep{littman1994markov}, defined as a tuple $\langle \mathcal{I}, \mathcal{S},\{\mathcal{O}_i\}_{i\in \mathcal{I}}, \{\mathcal{A}_i\}_{i\in \mathcal{I}}, \mathcal{T}, \{\mathcal{R}_i\}_{i\in \mathcal{I}}, \gamma \rangle$; $\mathcal{I}\!=\![1,...,n]$ is a set of $n$ agents, $\mathcal{S}$ is the state space, $\{\mathcal{O}_i\}_{i\in \mathcal{I}}$ is the set of observation spaces, $\{\mathcal{A}_i\}_{i\in \mathcal{I}}$ is the set of action spaces, $\mathcal{T}$ is the state transition function, $\{\mathcal{R}_{i}\}_{i\in\mathcal{I}}$ is the set of reward functions, and $\gamma$ is the discount factor. 
Each agent $i$ chooses an action $a_{i}$ according to its stochastic policy $\pi_{\theta_i}\!: \mathcal{O}_i \times \mathcal{A}_i \to [0, 1]$, where $\theta_i$ denotes agent $i$'s policy parameters. 
Then, the joint action $a\!=\!\{a_{i},a_{-i}\}$ yields a transition to the next state according to the state transition function $\mathcal{T}\!: \mathcal{S}\times \{\mathcal{A}_i\}_{i\in\mathcal{I}} \to \mathcal{S}$. 
Note that the notation $-i$ indicates all other agents except agent $i$.
Then, agent $i$ obtains a reward as a function of the state and the joint action $r_i\!: \mathcal{S} \times \{\mathcal{A}_i\}_{i\in \mathcal{I}} \to \mathbb{R}$, and receives its private observation according to $o_i\!: \mathcal{S}\to \mathcal{O}_i$. 
Each agent aims to maximize its own total expected discounted return $R_i\!=\!\mathbb{E}_{\pi}[\sum_{t=1}^T\gamma^t r_{i}^t]$, where $r_{i}^{t}$ denotes $i$'s reward received at timestep $t$, $\pi$ denotes the joint policy, and $T$ denotes the episodic horizon. 

\subsection{Multiagent deep deterministic policy gradient}\label{section:maddpg}
To stabilize learning in MARL, MADDPG~\cite{lowe2017multi} introduced the centralized training and decentralized execution paradigm, in which the centralized critic conditions on the global information and the decentralized actor only depends on the agent's local observation. 
Specifically, a centralized critic for agent $i$ is defined as $Q_i^{\mu}(o,a_{i},a_{-i})\!=\!\mathbb{E}_{\mu}[R_i|o^1\!=\!o, a^1\!=\!\{a_{i},a_{-i}\}]$, where $o$ and $\mu$ denote the joint observation and policy, respectively. 
The policy gradient for agent $i$'s deterministic policy $\mu_{\theta_i}$ (abbreviated as $\mu_{i}$) with respect to the expected return $J(\theta_i)\!=\!\mathbb{E}_{\mu}[R_i]$ is:
\bea\label{eq:gradient}
\begin{split}
&\nabla_{\theta_i}J(\theta_i)\\
&=\mathbb{E}_{o,a\sim \mathcal{D}}[\nabla_{\theta_i}\mu_i(a_i|o_i)\nabla_{a_i}Q_i^{\mu}(o, a_{i}, a_{-i})|_{a_i=\mu_i(o_i)}],
\end{split}
\eea
where $\mathcal{D}$ denotes the replay buffer. The buffer $\mathcal{D}$ stores the tuples $(o,o',a,r)$ where $o'$ is the next joint observation and $r$ is the joint reward. 
The centralized critic $Q_i^{\mu}$ is updated by minimizing the following loss function:
\bea\label{eq:critic_loss}
\begin{split}
\mathcal{L}(\theta_i) &= \mathbb{E}_{o,o',a,r\sim \mathcal{D}}[Q_i^{\mu}(o, a_{i},a_{-i}) - y]^2,\\
\text{s.t.}\quad y &= r_i + \gamma Q_i^{\mu'}(o', a'_{i},a'_{-i})|_{a'_j=\mu'_j(o'_j),\forall j\in \mathcal{I}},
\end{split}
\eea
where $\mu'\!=\!\{\mu_{\theta'_i}\}_{i\in\mathcal{I}}$ denotes the set of target policies.

\subsection{Minimax MultiAgent deep deterministic policy gradient}\label{section:m3ddpg}
To learn a robust policy in MARL, an agent should account for the changes in the policy of other simultaneously learning agents. 
\cite{li2019robust} proposed M3DDPG, a robust multiagent learning approach based on the minimax optimization by assuming other agents are adversarial agents.
Specifically, each agent $i$ in \cite{li2019robust} optimizes the following learning objective: 
\bea\label{eq: minimax_a}
\max_{\theta_i}\min_{a_{-i}\in \mathcal{B}_{-i}}Q_i^{\mu}(o,a_i, a_{-i})|_{a_i = \mu_i(o_i)},
\eea
where $\mathcal{B}_{-i}$ is a compact constraint set of $a_{-i}$ (e.g., a $l_p$ norm ball). 
Because solving a general nonconvex-nonconcave minimax optimization problem is generally intractable~\citep{daskalakis2020complexity}, M3DDPG replaces the inner minimization with a one-step gradient descent: 
\bea\label{eq:m1}
\begin{split}
a_{-i}^* &= \arg\min_{a_{-i}}Q_i^{\mu}(o,a_i, a_{-i}) \\
&\approx a_{-i} - \alpha_{-i}\nabla_{a_{-i}}{Q_i^{\mu}}(o,a_i, a_{-i}),
\end{split}
\eea
where $\alpha_{-i} \ge 0$ denotes the learning rate.
\section{Approach}\label{section:approach}
While the single-step gradient approximation in M3DDPG~\cite{li2019robust} (see Section \ref{section:m3ddpg}) improves robustness, we note that the framework has several limitations:
\begin{itemize}
    \item The single-step gradient approximation can explore the locally worst situation and thus still lead to unsatisfying behavior when testing with new opponents that have drastically different strategies. 
    Applying Equation~\eqref{eq:m1} multiple times for the inner minimization can potentially alleviate this issue, but this results in a double-loop approach in solving Equation~\eqref{eq: minimax_a}, which is computationally prohibitive \cite{nichol18reptile}. 
    \item Moreover, the one-step gradient descent approximation can only compute the upper bound of the inner minimization problem because the original problem cannot be solved to a global optimum. Hence, for the outer level, maximizing an upper bound of the inner objective cannot guarantee the maximum of the original objective in Equation \eqref{eq: minimax_a}. 
    In other words, even though one-step gradient descent approximation cannot find a perturbation that results in the smallest $Q_i^{\mu}$, such perturbation can exist. 
\end{itemize}
As we detail in this section, we address these issues by employing convex relaxation of neural networks and solving the inner minimization to explore the approximate \textit{globally} worst situation while achieving \textit{certified} robustness from the guaranteed bound of the convex relaxation.

\subsection{Convex relaxation of neural networks}\label{sec:convex-relaxation}
We propose to convexify the centralized action-value function in MARL and efficiently solve the inner minimization problem in Equation~\eqref{eq: minimax_a}. 
Specifically, we assume that $Q^{\mu}_{i}$ is parameterized by fully connected networks with $L$ layers with an input $x_0\!=\!(o,a_i, a_{-i})$. 
Then, $Q_i^{\mu}$ can be expressed by the following form:
\bea\label{eq:Q_MLP}
\begin{split}
z^l &= W_c^lx^{l-1}, x^l = \sigma(z^l), \forall l=1,..., L-1\\
Q_{i}^{\mu} &= W_c^{L}z^L,
\end{split}
\eea
where $\sigma(\cdot)$ denotes the nonlinear activation function, and $W_c^l$ is the weight at layer $l$. 
For clarity, we drop the bias terms without loss of generality. 
Due to the nonconvexity of the activation function, verifying the robustness property of $Q_{i}^{\mu}$ over a compact set of $x_0$ is difficult~\citep{wong2018provable,weng2018towards}. 
To address this, we employ a convex relaxation technique \cite{wong2018provable} to verify robustness verification of neural networks and apply the following linear convexification to the centralized Q-function by assuming the ReLU activation function:
\bea\label{eq:Q_MLP_relax}
\begin{split}
z^l &= W_c^lx^{l-1}, x^l \le u^{l} \odot (z^l - l^l) \oslash (u^l - l^l)\\
x^l &\ge 0, x^l \ge z^l, \forall l=1,\ldots, L-1\\
\bar{Q}^{\mu}_{i} &=  W_c^{L}z^L,
\end{split}
\eea
where $l^l$ and $u^l$ are lower and upper bounds for $z^l$, respectively, and $\odot$ and $\oslash$ are the element-wise multiplication and division, respectively. Note that $\bar{Q}_i^{\mu}$ indicates the relaxed version of ${Q}_i^{\mu}$ (i.e., $l^l \le z^l \le u^l$). 
Thanks to this relaxation, all equations in \eqref{eq:Q_MLP_relax} are convex, so $\min_{a_{-i}}\bar{Q}_i^{\mu}(o,a_i, a_{-i})$ is a linear programming and can be solved efficiently. In the evaluation, we empirically show that this new certification module is computationally efficient and does not add much burden on top of a base MARL algorithm.


\begin{figure}[t]
  \centering
  \includegraphics[width=0.55\linewidth]{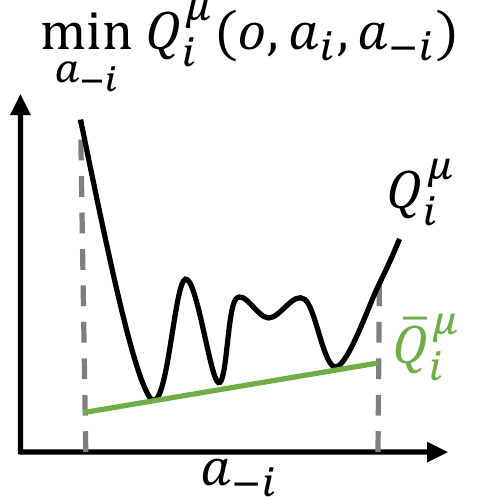}
  \caption{Illustration of the convex relaxation approach for solving the inner minimization in Equation~\eqref{eq: regularizer_a_relax}.}
  \label{fig:convexframework}
\end{figure}

\subsection{Solving minimiax optimization via convex relaxation}\label{subsec:minimax}
Here, we employ the convex relaxation technique discussed in \Cref{sec:convex-relaxation} to solve the inner minimization problem approximately. 
Specifically, we propose to replace the inner minimization problem in Equation~\eqref{eq: minimax_a} with the following relaxed objective:
\bea\label{eq: regularizer_a_relax}
\begin{split}
\bar{Q}_i^{\mu}(o,a_i, a^*_{-i}) &= \min_{a_{-i}\in\mathcal{B}_{-i}}\bar{Q}_i^{\mu}(o,a_i, a_{-i})\\
&\le\min_{a_{-i}\in\mathcal{B}_{-i}}Q_i^{\mu}(o,a_i, a_{-i}),
\end{split}
\eea
where $\bar{Q}_i^{\mu}(o,a_i, a_{-i})$ is the lower bound of $Q_i^{\mu}(o,a_i, a_{-i})$ from the relaxation in Equation~\eqref{eq:Q_MLP_relax} (see~\Cref{fig:convexframework}). $\bar{Q}_i^{\mu}(o,a_i, a_{-i})$ is also a function of $\mu$ as well and is a surrogate of the original non-convex nonlinear objective. 
The main advantage of the convex relaxation in Equation~\eqref{eq: regularizer_a_relax} over \eqref{eq:m1} is that the former does not need the step size hyperparameter $\alpha_{-i}$, {which can be difficult to tune}. 
The performance of \cite{li2019robust} is highly sensitive to the step size, and it is difficult to tune. 
By contrast, our convex relaxation problem can be efficiently solved without needing the step size. 
With this lower bound, we can reformulate the outer maximization problem as:
\bea\label{eq: minimax_a_relax}
&&\max_{\theta_i}  \big[ (1-\kappa_i)Q_i^{\mu}(o,a_i, a_{-i}) + \kappa_i \bar{Q}_i^{\mu}(o,a_i, a^*_{-i})\big]\nonumber\\
&&\text{with}\ \  {a_i = \mu_i(o_i)},
\eea
where $0\!\le \kappa_i\!\le 1$ is a weight coefficient for the term that accounts for the policy change of the other agents. 
Because we maximize the lower bound of the inner minimization problem, the original inner objective is guaranteed to be maximized. 
Such a guarantee provides robustness certificates for agent $i$ as it considers the worst-case scenarios caused by other learning agents. 
By setting $\kappa_i\!\ne\!1$,  we do not entirely use the relaxed inner objective (i.e., $\bar{Q}_i^{\mu}(o,a_i, a^*_{-i})$) as the objective of the outer maximization problem for the sake of training stability, as a relaxation gap might be big especially in the early training process. 
Instead, a combination of the original objective and its relaxed one is used as the objective for the outer maximization problem, as shown in Equation~\eqref{eq: minimax_a_relax}. 
Because this inner minimization needs to be solved whenever the policy is updated, the convex relaxation problem in Equation~\eqref{eq: regularizer_a_relax} should be efficient enough with a tight bound. Therefore, there is a trade-off to choose a certain convex relaxation method among many candidates, in which we refer to the appendix for details.

\begin{algorithm}[t]
\caption{Robust MARL via convex relaxation (ROMAX)}
\label{alg:minimax}
\small
\begin{algorithmic}[1]
\State \textbf{Require}: batch size $S$, actor learning rate $\alpha_a$, critic learning rate $\alpha_c$, target update $\tau$, random process $\mathcal{N}$, episode length $T$
\State Initialize replay buffer $\mathcal{D}$
\For {Episode=$1\ldots$}
    \State Initialize environment and get initial observations $o$
    \For {$t=1...{T}$}
        \State For each agent, select action $a_i=\mu_{\theta_i}(o_i) + \mathcal{N}_t$
        \State Execute joint action $a$ and receive $r$ and $o'$
        \State Store $(o,a,r,o')$ into $\mathcal{D}$, set $o\leftarrow o'$
        \For {Each agent $i\in\mathcal{I}$}
            \State Get $S$ samples $(o,a,r,o')$ from $\mathcal{D}$
            \State Solve inner optimization via relaxation in \eqref{eq:critic_loss_minimax}
            \State Update critic via loss function in \eqref{eq:critic_loss_minimax} with $\alpha_c$
            \State Solve inner optimziation via relaxation in \eqref{eq:gradient_minimax}
            \State Update actor via policy gradient in \eqref{eq:gradient_minimax} with $\alpha_a$
        \EndFor
        \State Update target network $\theta_i =  \tau \theta_i + (1-\tau)\theta'_i$
    \EndFor
\EndFor
\end{algorithmic}
\end{algorithm}

\subsection{Integrating with MARL algorithm}

Our framework based on convex relaxation in Section \ref{subsec:minimax} can be readily integrated into general MARL frameworks. 
We implement our method based on MADDPG (see Section \ref{section:maddpg}).
Integrating the minimax formulation and the convex relaxation in Equation~\eqref{eq: minimax_a_relax} together with the actor update in Equation~\eqref{eq:gradient} yields:
\bea\label{eq:gradient_minimax}
\begin{split}
\nabla_{\theta_i}J(\theta_i) &= \mathbb{E}_{o,a\sim \mathcal{D}}\Big[\nabla_{\theta_i}\mu_i(o_i)\nabla_{a_i}\Big(\kappa_i \bar{Q}_i^{\mu}(o,a_i, a^*_{-i}) \\
&\ \ \ \ \  + (1-\kappa_i)Q_i^{\mu}(o,a_i, a_{-i})\Big)\Big], \\
a_i &= \mu_i(o_i),\quad a^*_{-i} =\arg \min_{a_{-i}\in\mathcal{B}_{-i}}\bar{Q}_i^{\mu}(o,a_i, a_{-i}),
\end{split}
\eea
where $\mathcal{B}_{-i} = \mathcal{B}_{-i}((a_j=\mu_j(o_j),\forall j\ne i), \epsilon)$ is a $l_p$ ball centered at $(a_j=\mu_j(o_j),\forall j\ne i)$ with a radius $\epsilon$. 
Then, the critic is updated by:
\bea\label{eq:critic_loss_minimax}
\begin{split}
\mathcal{L}(\theta_i) &= \mathbb{E}_{o,o',a,r\sim \mathcal{D}}[Q_i^{\mu}(o, a_i,a_{-i}) - y]^2,\\
y &= r_i + \gamma \Big((1-\kappa_i) Q_i^{\mu'}(o', a'_i,a'_{-i}) \\
&\ \ \ \ \ \ \ + \kappa_i \bar{Q}_i^{\mu'}(o',a'_i, a'^*_{-i})\Big)\\
a'_i&=\mu'_i(o'_i), \quad a'^*_{-i} = \arg \min_{a'_{-i}\in\mathcal{B}'_{-i}}\bar{Q}_i^{\mu'}(o',a'_i, a'_{-i}),
\end{split}
\eea
where $\mathcal{B}'_{-i} = \mathcal{B}'_{-i}((a'_j=\mu'_j(o'_j),\forall j\ne i), \epsilon')$ is a $l_p$ ball centered at $(a'_j=\mu'_j(o'_j),\forall j\ne i)$ with a radius $\epsilon'$. We summarize our algorithm in Algorithm \ref{alg:minimax}. 
\begin{figure}[t]
    \centering
    \includegraphics[width=0.9\linewidth]{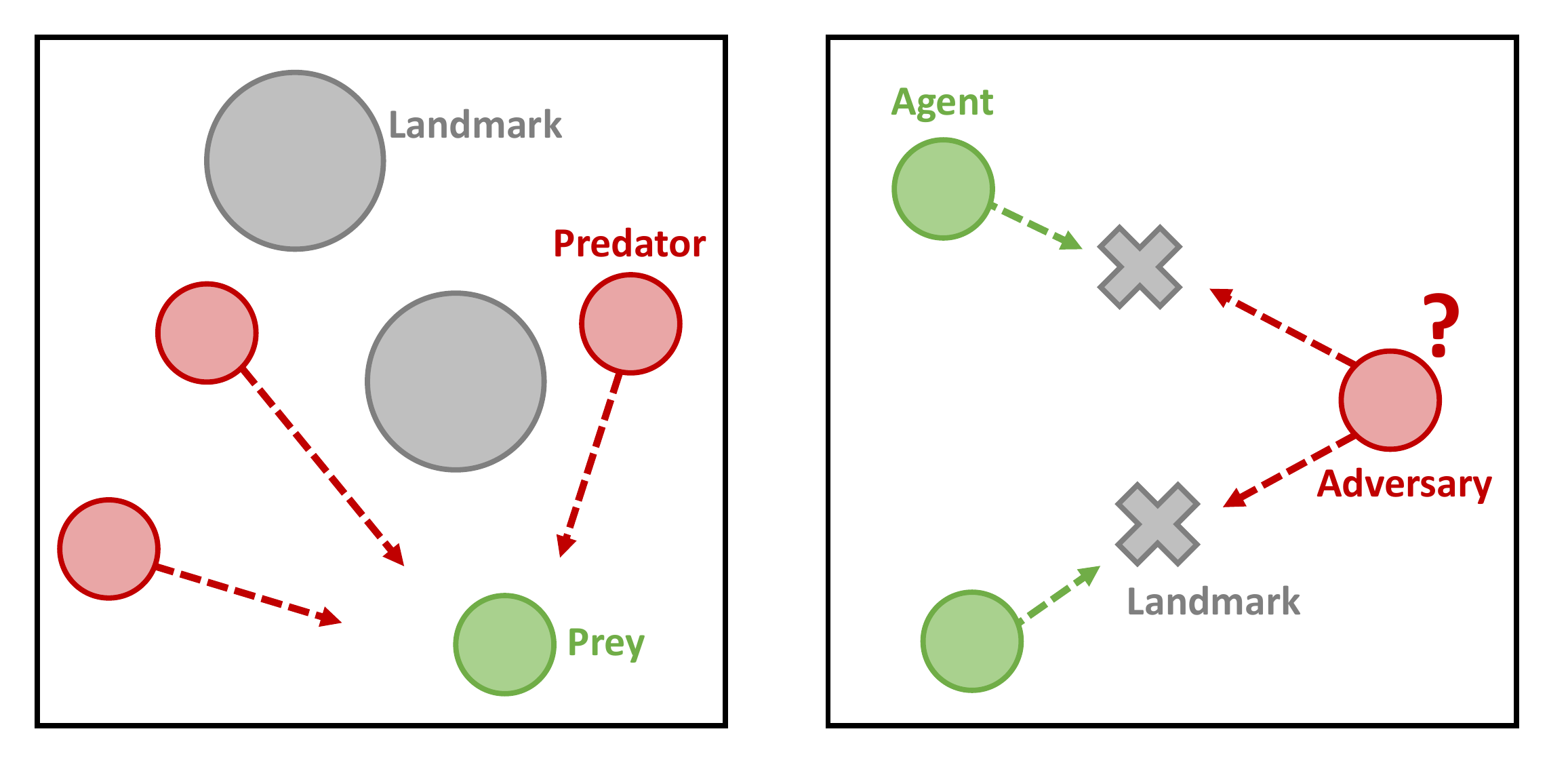}
    \caption{Illustration of the considered tasks: Predator-prey (Left) and physical deception (Right); Reproduced from~\cite{li2019robust}.}
    \label{fig:illustration_tasks}
\end{figure}

\section{Experiments}\label{section:results}
\subsection{Evaluation domains and baselines}
We evaluate our approach in mixed cooperative-competitive tasks from the multiagent particle benchmark \cite{lowe2017multi}. 
In these tasks, there are $n_c$ cooperative agents, $n_a$ adversary agents, and $n_L$ landmarks in a 2D continuous space. 
We focus on tasks that include adversarial agents such that agents need to have diverse strategies to outperform opponents and thus robustness is an important factor. 
Below are some descriptions of considered tasks with illustration in Figure \ref{fig:illustration_tasks}.

\begin{table*}[t]
\begin{center}
\begin{tabular}{c||c|c|c||c}
\hline\\[-1em]
\backslashbox{\red{Adv}}{\blue{Agent}}       &   MADDPG     & \makecell{M3DDPG} & ROMAX &  ${R}_{\text{\red{Adv}}}$    \\
\\[-1em]\hline\hline
\makecell{MADDPG}      & $\Big(\makecell{\red{-0.017}{\scriptscriptstyle \pm 0.012}, \\ \blue{-0.550}{\scriptscriptstyle \pm 0.017 }}\Big)$    &$\Big(\makecell{\red{0.160}{\scriptscriptstyle \pm 0.045}, \\ \blue{-0.502}{\scriptscriptstyle \pm 0.053}}\Big)$   &$\Big(\makecell{\red{0.031}{\scriptscriptstyle \pm 0.020}, \\ \blue{-0.406}{\scriptscriptstyle \pm 0.025}}\Big)$  & $0.174 \scriptscriptstyle \pm 0.080 $ \\
\hline
\makecell{M3DDPG}      & $\Big(\makecell{\red{0.307}{\scriptscriptstyle \pm 0.043},\\  \blue{-0.718}{\scriptscriptstyle \pm 0.051}}\Big)$    & $\Big(\makecell{\red{0.250}{\scriptscriptstyle \pm 0.048},\\  \blue{-0.609}{\scriptscriptstyle \pm 0.060}}\Big)$  &$\Big(\makecell{\red{-0.043}{\scriptscriptstyle \pm 0.031},\\  \blue{-0.290}{\scriptscriptstyle \pm 0.042}}\Big)$  & $0.514\scriptscriptstyle \pm 0.158$ \\
 \hline
ROMAX  &  $\Big(\makecell{\red{0.560}{\scriptscriptstyle \pm 0.032},\\  \blue{-1.093}{\scriptscriptstyle \pm 0.037}}\Big)$   & $\Big(\makecell{\red{0.428}{\scriptscriptstyle \pm 0.055},\\  \blue{-0.936}{\scriptscriptstyle \pm 0.057}}\Big)$  &$\Big(\makecell{\red{0.132}{\scriptscriptstyle \pm 0.020},\\  \blue{-0.477}{\scriptscriptstyle \pm 0.026}}\Big)$  &  $1.12\scriptscriptstyle \pm 0.183$\\
\hline
${R}_{\text{\blue{Agent}}}$ & $-2.361\scriptscriptstyle \pm 0.230$     & $-2.047\scriptscriptstyle \pm 0.193$ & $-1.173\scriptscriptstyle \pm 0.083$ &  \\
\hline
\Xhline{2\arrayrulewidth}
\Xhline{2\arrayrulewidth}
 & MADDPG & \makecell{M3DDPG} & ROMAX     & \\
\hline

\makecell{${R}_{\text{overall}}$ \\ $ = {R}_{\text{\red{Adv}}} + {R}_{\text{\blue{Agent}}}$} & $-2.187$ & $-1.533$& \bm{$-0.053$} & \\
\hline
\end{tabular}
\end{center}
\caption{
Evaluation in the predator-prey task. Predator and prey correspond to adversary (Adv for short) and agent in the table, respectively. Each pair is evaluated for $250$ episodes, i.e., $10$ episodes for each of the $5\!\times\!5\!=\!25$ pairs of random seeds. $(\red{\bullet},\blue{\bullet})$ in each cell denotes the mean/standard error of the reward per step in the episode of the adversaries and agents, respectively. The higher the return is, the better the policy is. For each column, different adversaries compete against the same agent, and a higher adversary reward indicates better performance against the same agent; row-wise for the agents. In the last row, we summarize the overall robustness results for playing both teams via the metric ${R}_{\text{overall}}$.}
\label{table:predator}
\end{table*}

\begin{table*}[t]
\begin{center}
\begin{tabular}{c||c|c|c||c}
\hline\\[-1em]
\backslashbox{\red{Adv}}{\blue{Agent}}       &   MADDPG     & \makecell{M3DDPG} & ROMAX  &  ${R}_{\text{\red{Adv}}}$    \\
\\[-1em]\hline\hline
\makecell{MADDPG}      & $\Big(\makecell{\red{-0.795}{\scriptscriptstyle \pm 0.017}, \\ \blue{0.482}{\scriptscriptstyle \pm  0.005}}\Big)$    &$\Big(\makecell{\red{-0.689}{\scriptscriptstyle \pm 0.031}, \\ \blue{0.248}{\scriptscriptstyle \pm 0.020}}\Big)$   &$\Big(\makecell{\red{-0.814}{\scriptscriptstyle \pm 0.032}, \\ \blue{0.338}{\scriptscriptstyle \pm 0.0199}}\Big)$  &  $-2.298\scriptscriptstyle \pm  0.061$\\
\hline
\makecell{M3DDPG}      & $\Big(\makecell{\red{-0.742}{\scriptscriptstyle \pm 0.029},\\  \blue{0.225}{\scriptscriptstyle \pm 0.021}}\Big)$    & $\Big(\makecell{\red{-0.819}{\scriptscriptstyle \pm 0.018},\\  \blue{0.467}{\scriptscriptstyle \pm 0.004}}\Big)$  &$\Big(\makecell{\red{-0.839}{\scriptscriptstyle \pm 0.037},\\  \blue{0.271}{\scriptscriptstyle \pm 0.020}}\Big)$  &  $-2.4\scriptscriptstyle \pm  0.050$ \\
 \hline
ROMAX  &  $\Big(\makecell{\red{-0.572}{\scriptscriptstyle \pm 0.0282},\\  \blue{0.128}{\scriptscriptstyle \pm 0.019}}\Big)$   & $\Big(\makecell{\red{-0.613}{\scriptscriptstyle \pm 0.033},\\  \blue{0.133}{\scriptscriptstyle \pm 0.0193}}\Big)$  &$\Big(\makecell{\red{-0.512}{\scriptscriptstyle \pm 0.010},\\  \blue{0.283}{\scriptscriptstyle \pm 0.003}}\Big)$  &  $-1.697\scriptscriptstyle \pm  0.048$\\
\hline
${R}_{\text{\blue{Agent}}}$ & $0.835\scriptscriptstyle \pm 0.150 $     & $0.848\scriptscriptstyle \pm  0.139$ & {$0.892\scriptscriptstyle \pm  0.033$} &  \\
\hline
\Xhline{2\arrayrulewidth}
\Xhline{2\arrayrulewidth}
 & MADDPG & \makecell{M3DDPG} & ROMAX    & \\
\hline
\makecell{${R}_{\text{overall}}$ \\ $ = {R}_{\text{\red{Adv}}} + {R}_{\text{\blue{Agent}}}$} & $-1.463$ & $-1.552$& \bm{$-0.805$}&\\
\hline
\end{tabular}
\end{center}
\caption{Evaluation in the physical deception task. The evaluation settings and metrics shown in this table are the same as those in Table 1.}
\label{table:deception}
\end{table*}
 
\begin{itemize}
\item \textbf{Predator-prey.} $n_a\!=\!3$ slower cooperative predators aim to catch the $n_c\!=\!1$ faster prey. $n_L\!=\!2$ landmarks are unmovable and can impede the way of all of the agents. Once there is a collision between predators and the prey, the former get rewarded while the latter gets penalized.
\item \textbf{Physical deception.} There are $n_a\!=\!1$ adversary and $n_c\!=\!2$ agents together with $n_L\!=\!2$ landmarks in the environments. The adversary aims to occupy a target landmark without knowing which one of the two landmarks is the target. As a result, agents must split and cover all landmarks to deceive the adversary.
\end{itemize}

\textbf{Transfer to real robot learning.} 
We note that these tasks closely coincide with real-world robotic missions. For the predator-prey, multiple robots can be deployed to chase an intelligent moving target (e.g., an intruder in a market). For physical deception, we can deploy robots to protect assets of interest with intelligent behaviors to deceive opponents.
The fidelity of the models and perception required in simulation can be achieved in the real world via sensors such as cameras, velocity meters, and LiDAR. 
Sim-to-real is known to be difficult, because the behaviors of other agents deployed in the environment in the real-world may differ significantly from the simulation (e.g., due to varying transition dynamics). 
This is exactly what this work aims to address: the certified and improved robustness will enhance the resilience and applicability of multiagent algorithms from sim-to-real.
Lastly, the learned policy can be easily transferred on-board, and generated actions can be further executed by a lower-level controller if necessary.

\textbf{Baselines.} We compare ROMAX to M3DDPG~\cite{li2019robust}, a robust MARL algorithm that also applies the minimax formulation but solves the inner optimization approximately via the one-step gradient descent. 
We also compare our algorithm to MADDPG~\cite{lowe2017multi}, which uses the centralized critic but does not solve minimax.
Implementation details and hyperparameters are specified in the appendix. 


\subsection{Results}
\textbf{Question 1:} \textit{How much does ROMAX improve the robustness of trained policies?}

To answer this question and test robustness, each policy from one team is evaluated against a diverse set of policies from the other team. Then the adversaries' policies trained by one algorithm under each random seed will be evaluated against the agents' policy trained by all of the \textit{other} algorithms under \textit{all} random seeds; vice-versa for the agent's policy. 

As Table 1 and 2 demonstrate, for both tasks, ROMAX can train more robust policies for both teams in a competitive game. For each adversary, when competing against the same set of diverse agents, our adversary get the highest return; see the $R_{\text{\red{Adv}}}$ columns in the tables. Similar conclusion can be made for the agents given the $R_{\text{\blue{Agent}}}$ rows in the tables. These results demonstrate that, via computing the approximate global worst-case situation, policies can generalize and perform well when tested against unseen peer agents' policies. We also note that M3DDPG is outperformed by MADDPG in Table 2 (see the overall robustness results). 
This might be due to the sensitive step-size parameter of M3DDPG in Equation~\eqref{eq:m1}. This observation implies that a tuned step size for one task cannot generalize to another one and also shows the advantage of ROMAX. {Regarding the computation efficiency, we empirically observe that the factor between wall-clock time per iteration of ROMAX (with certification) and that of MADDPG (without certification), is close to 1 (i.e., $1.08$, averaged among multiple seeds). This validates that our certification module is computationally efficient.}

\textbf{Question 2:} \textit{How much can disruptive policies exploit a fixed robust policy?}

\begin{figure}[t]
\begin{center}\label{fig:predator_prey}
\includegraphics[scale=0.4]{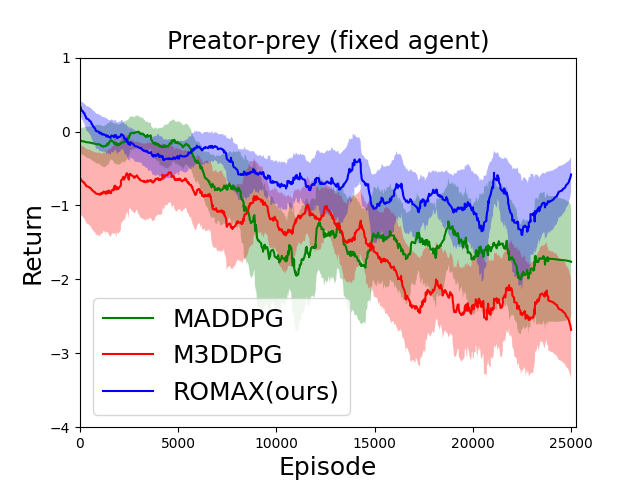}
\caption{
The performance of the fixed agent (prey) during the training of disruptive adversaries (predators) with $3$ seeds. Higher return implies a more robust policy.
}
\end{center}
\end{figure}

To answer this question, we construct a disruptive policy in the predator-prey task by 1) training both teams with each algorithm until convergence, 2) fixing the prey policy, and 3) training new adversary predators policies based on MADDPG that attempt to exploit this fixed prey trained by each method. 
In Figure 3, the robustness results of the fixed prey trained with different algorithms are shown. 
As the disruptive adversaries' training proceeds, the fixed prey's return decreases as expected. 
However, ROMAX achieves the highest return compared to other methods, validating the robustness advantage of our approach. 
We observe that M3DDPG and MADDPG perform similarly in this analysis, possibly due to the sensitive tuning of the step size.
\section{Conclusion}\label{section:conclusion}
In this paper, we propose a robust reinforcement learning algorithm for a multirobot system. 
To robustify learning, we consider the learning of other agents based on the worst-case scenario criterion, which inherently leads to a minimax formulation. 
As minimax formulation is computationally expensive to solve, convex relaxation of neural networks is applied to solve the inner minimization problem. 
By convex relaxation, agents can account for peer agents that possibly have drastically different behaviors, and a certified bound of the original optimization problem can be gained. 
We believe this is the first work that integrates robustness verification in MARL. 
Our algorithm outperforms existing robust MARL algorithms in mixed cooperative-competitive tasks. 

There are a few important directions for future works. First, we would like to develop tight but efficient convex relaxation-based methods for neural network robustness verification.
Moreover, there are several real-world robustness applications, including observation perturbation, actuation fault, malicious/stealthy attack, communication delay, that we would like to test our approach on.
Lastly, developing principled and general learning methods with theoretical guarantees (e.g., convergence analysis) will be a meaningful direction. 

\appendix
\section{Appendix}\label{section:appendix}
\subsection{Repositories}\label{section:specs}
The multiagent particle environments we used in simulation is from \url{https://github.com/openai/multiagent-particle-envs}. We use the implementation of the base algorithm MADDPG from \url{https://github.com/shariqiqbal2810/maddpg-pytorch}. 
Note that with relaxations in Equation \eqref{eq:Q_MLP_relax}, $\bar{Q}_i^{\mu}(o,a_i, a^*_{-i})$ is no longer an explicit function with respect to its input without constraints. Then the framework {auto\_LiRPA} (\url{https://github.com/KaidiXu/auto\_LiRPA},~\cite{xu2020automatic}) is used to get $\bar{Q}_i^{\mu}(o,a_i, a^*_{-i})$ efficiently and automatically. 

\subsection{Choice of convex relaxation methods}
For robustness verification of neural networks there are many convex relaxation based methods, from which we need to choose one for Equation \eqref{eq: regularizer_a_relax}. 
When there is a trade-off to choose a certain convex relaxation method among many candidates, we can get $\bar{Q}_i^{\mu}(o,a_i, a^*_{-i})$ as a convex combination of the bounds from different methods~\citep{zhang2019towards}. For example, Interval Bound Propagation (IBP,~\citep{gowal2018effectiveness}) and CROWN-IBP~\citep{zhang2019towards} have their respective strengths and shortcomings in terms of bound tightness, sensitivity to hyper-parameters, computational cost with the training going on. As a result, we can have:
\bea\label{eq:two_bounds}
\begin{split}
\bar{Q}_i^{\mu}(o,a_i, a^*_{-i}) &= \beta\bar{Q}_{i,\text{IBP}}^{\mu}(o,a_i, a^*_{-i}) \\
&+(1-\beta)\bar{Q}_{i,\text{CROWN-IBP}}^{\mu}(o,a_i, a^*_{-i}),
\end{split}
\eea
with $\beta\in [0, 1]$ a tunable parameter which can change with the training iteration index increasing. As both $\bar{Q}_{i,\text{IBP}}^{\mu}(o,a_i, a^*_{-i})$ and $\bar{Q}_{i,\text{CROWN-IBP}}^{\mu}(o,a_i, a^*_{-i})$ are the lower bounds of $Q_i^{\mu}(o,a_i, a_{-i})$, so are their convex combination $\bar{Q}_i^{\mu}(o,a_i, a^*_{-i})$. Hence, the property of certified robustness is kept. 

\subsection{Hyperparameter}
Some key hyperparameters are shown in Table \ref{table:Hyperparameters}.
\begin{table}[hbt!]
\begin{center}
\begin{tabular}{c|c||c|c}
\hline\\[-1em]
Episode length & $25$   & batch size & $1024$ \\
NN hidden dim  & $64$      & $\tau$ & $0.01$\\
learning rate & $0.01$    & $\epsilon_{\max}$ & $0.1$ \\
$\beta_{\min}$& $0.9$     & $\gamma$ & $0.99$ \\
\hline
\end{tabular}
\end{center}
\caption{Hyperparameters choices in the implementation.}
\label{table:Hyperparameters}
\end{table}





\section*{Acknowledgements}
Research supported by Scientific Systems Company, Inc. under research agreement \# SC-1661-04 and ARL DCIST under Cooperative Agreement Number W911NF-17-2-0181. Dong-Ki Kim was supported by IBM, Samsung (as part of the MIT-IBM Watson
AI Lab initiative), and Kwanjeong Educational Foundation Fellowship. 
We thank Amazon Web services for computational support.

\newpage
\bibliographystyle{IEEEtran}
\bibliography{example}

\balance


\end{document}